\documentclass{article} 
\usepackage{iclr2023_conference,times}

\usepackage{amsmath,amsfonts,bm}

\def\eqref#1{equation~\ref{#1}}

\def\1{\bm{1}}

\DeclareMathAlphabet{\mathsfit}{\encodingdefault}{\sfdefault}{m}{sl}
\SetMathAlphabet{\mathsfit}{bold}{\encodingdefault}{\sfdefault}{bx}{n}

\usepackage{hyperref}
\usepackage{url}
\usepackage{graphicx}
\usepackage[utf8]{inputenc}
\usepackage[T1]{fontenc}
\usepackage{booktabs}
\usepackage{amsfonts}
\usepackage{nicefrac}
\usepackage{microtype}
\usepackage{xcolor}
\usepackage{lipsum}
\usepackage{caption}
\usepackage{multirow}
\usepackage{listings} 
\newcommand*\samethanks[1][\value{footnote}]{\footnotemark[#1]}

\title{SteloCoder: a Decoder-Only LLM for Multi-Language to Python Code Translation}

\iclrfinalcopy

\author{%
  Jialing Pan$^{1,4}\thanks{Equal contribution and shared co-first authorship.}$
  \quad Adrien Sadé$^{2,3,4}\samethanks$
  \quad Jin Kim$^{4}$
  \quad Eric Soriano$^{4}$
  \quad Guillem Sole$^{4}$
  \quad Sylvain Flamant$^{4}$ \\ 
  $^1$ Stanford University\\
  $^2$ University of California, Berkeley\\
  $^3$ École des Ponts ParisTech\\
  $^4$ Esperanto Technologies Inc.\\
  \texttt{jlpan@stanford.edu}\\
   \texttt{\{adrien.sade, jin.kim\}@esperantotech.com} \\
   \texttt{\{eric.soriano, guillem.sole, sylvain.flamant\}@esperantotech.com}
 }

\begin{document}

\maketitle

\begin{abstract}
With the recent focus on Large Language Models (LLMs), both StarCoder \citep{li2023starcoder} and Code Llama \citep{rozière2023code} have demonstrated remarkable performance on code generation. However, there is still a need for improvement in code translation functionality with efficient training techniques. In response to this, we introduce SteloCoder, a decoder-only StarCoder-based LLM designed specifically for multi-programming language-to-Python code translation. In particular, SteloCoder achieves C++, C\#, JavaScript, Java, or PHP-to-Python code translation without specifying the input programming language. We modified StarCoder's model architecture by incorporating a Mixture-of-Experts (MoE) technique featuring five experts and a gating network for multi-task handling.  Experts are obtained by StarCoder fine-tuning. Specifically, we use a Low-Rank Adaptive Method (LoRA) technique, limiting each expert size as only 0.06\% of number of StarCoder's parameters. At the same time, to enhance training efficiency in terms of time, we adapt curriculum learning strategy and use self-instruct data for efficient fine-tuning. As a result, each expert takes only 6 hours to train on one single 80Gb A100 HBM. With experiments on XLCoST datasets, SteloCoder achieves an average of 73.76 CodeBLEU score in multi-programming language-to-Python translation, surpassing the top performance from the leaderboard by at least 3.5. This accomplishment is attributed to only 45M extra parameters with StarCoder (15.5B) as the backbone and 32 hours of valid training on one 80GB A100 HBM. The source code is released here: \href{https://github.com/sade-adrien/SteloCoder}{https://github.com/sade-adrien/SteloCoder}.

\end{abstract}

\section{Introduction}

The adoption of LLMs, such as  GPT \citep{brown2020language}, LLaMA \citep{touvron2023llama}, and StarCoder \citep{li2023starcoder}, is on the rise. These models were all pre-trained based on broad range and great amount of data, establishing foundation models that have the capability to adapt into specific downstream tasks. Significant progress has been made in the field of code generation such as Natural Language (NL) to Programming Language (PL) conversion due to its potential to enhance productivity. Notably, language models such as Code Llama \citep{rozière2023code}, StarCoder \citep{li2023starcoder}, and CodeTrans \citep{elnaggar2021codetrans} can perform code generation (NL to PL) and code summarization (PL to NL) of high reliability. Moreover, the application of language models in code translation between programming languages (PL to PL), such as COBOL to Python, has already shown remarkable efficiency \citep{kulshrestha2022cobol2vec}. With an encoder-decoder structure, CodeT5 \cite{wang2021codet5} established its outstanding performance in code translation. 

Over the first years following the release of the Transformer architecture \citep{vaswani2023attention}, the majority of research in code translation was predominantly focused on the encoder-decoder architecture \citep{yang2023harnessing}. Specifically, CodeT5 \citep{wang2021codet5} relies on an encoder-decoder architecture for code understanding and generation. However, more recent decoder-only models gained in popularity with the introduction of highly performing generation models \citep{brown2020language} \citep{chowdhery2022palm}
\citep{touvron2023llama}.

In this paper, we present SteloCoder, a decoder-only StarCoder-based LLM specialized in translating code from five PLs (C++, C\#, PHP, JavaScript, Java) to Python. We propose to optimize the training stage by utilizing the following strategy:

\begin{itemize}
  \item We use curriculum learning (CL) \citep{10.1145/1553374.1553380} to achieve one-PL-to-Python code translation. Specifically, we train from simple tasks progressively to more challenging ones. StarCoder would first be fine-tuned on code snippet-type data and subsequently on code program-type data. 
  \item We utilize LoRA (Low-Rank Adaptation) \citep{hu2021lora} technique to reduce trained parameters and be memory efficient. After fine-tuning, only LoRA weights were preserved, and each set serves as one expert for a specific PL-to-Python downstream task. One set only contributes for 0.06\% of the total model parameters. 
  \item SteloCoder incorporates a Mixture-of-Experts (MoE) \citep{shazeer2017outrageously} approach for routing. With a gating network, SteloCoder is able to recognize the input PL, route to the correct expert, and handle its translation downstream task effectively.
\end{itemize}

These strategies collectively enable SteloCoder to excel in code translation, particularly from diverse PLs to Python, while maintaining memory and time efficiency in the training process.

SteloCoder was tested on XLCoST datasets and achieved an average of 73.76 CodeBLEU score, surpassing the top performance by at least 3.5. This achievement is accomplished by only an addition of 45M parameters more than StarCoder's and a 32-hour training process on one 80 Gb A100 HBM.

\section{Related Work}
\paragraph{Pre-Training for Natural Language Generation} Since the introduction of the transformer and the attention mechanism \citep{vaswani2023attention}, the Natural Language Understanding (NLU) and Generation (NLG) fields have been addressed by a large part of the machine learning community resulting in a plethora of models released. The global architecture has remained the same over the years, but the improvement of data quality allowed continuous progress. Models like BART \citep{lewis2019bart} and T5 \citep{raffel2023exploring} reached state-of-the-art results in both understanding and generation tasks thanks to their encoder-decoder structures. While models like BERT (encoder-only) \citep{devlin2019bert} and GPT (decoder-only) \citep{brown2020language} specialize respectively in language understanding and language generation. 

\paragraph{Pre-Training for Programming Language Generation}  More recently, the scientific community has also been taking an interest in providing tools to work with PLs. Specifically, encoder-decoder models being adapted, and sometimes being fine-tuned, from pre-trained NL models have shown excellent result in code understanding and generation. For example, CodeBERT \citep{feng2020codebert}, inspired from BERT methodology, is one of the very first PL models and is able to generate text from multiple PLs. PLBART \citep{ahmad2021unified}, which re-uses BART architecture, thrives in code summarization, generation and translation. CodeT5 \citep{wang2021codet5}, which is established based on T5 architecture, provides state-of-the-art results in both code understanding and generation on various tasks. Those three models are general-purpose Programming Language Models (PLMs) which are able to perform a wide range of code understanding and generation tasks. Their evolving performance is mostly due to a growing quality in the data and training methods, along with punctual changes in the architecture.

\paragraph{Open Source Pre-Trained Models}
Since the release of LLaMA \citep{touvron2023llama}, the open source community has been extremely active, developing and fine-tuning LLMs every other day. This has allowed major progress and the release of expert models in plenty of fields. StarCoder is one example of this movement: it is the result of a collaborative work. It is based on GPT BigCode architecture (from SantaCoder work \citep{allal2023santacoder}), pre-trained on open source GitHub data (StarCoderBase), then fine-tuned to be Python-specific (StarCoder) \citep{li2023starcoder}. When released, it outperformed every other Python PLMs. The weights are openly released, and it is used as our base model.

\paragraph{Task-specific Experts} \cite{gururangan2021demix} introduces an expert mixture in the transformer block. Their model uses feed-forward networks experts as in the original MoE paper \citep{shazeer2017outrageously}.\cite{gao2022parameterefficient} proposed a parameter-efficient MoE architecture, using the Matrix Product Operator, a tensor decomposition method. This allows them to share parameters throughout the experts as they hypothesized that core parameters of the decomposition are the same for all experts. Our idea relies on the same grounds as both these papers, but we leverage the LoRA \citep{hu2021lora} matrices as our experts to be parameter-efficient. The backbone model is not impacted, and the experts are small rank matrices created by LoRA fine-tuning method. This contains the number of added parameters. Additionally, our routing function is similar as \cite{fedus2022switch}'s routing method. They restrain their model to choose only one expert and use the weighted experts' outputs sum to maintain differentiability.

\section{Approach}
The task of translating a language to another was the original reason that the transformer architecture was introduced \citep{vaswani2023attention}. It has been extensively studied before that \citep{cho2014properties} and ever since \citep{raffel2023exploring}. The common agreement is that the best performing models as of yet are encoder-decoder transformer-based architecture thanks to their natural inclination to both language understanding and generation.

Following the path, we aim to build and train our own PLM model for code translation. However, this process can be extremely long and expensive since best performing LLMs tend to be billions of parameters sized and tend to be trained on substantial amount of data. Training these models from scratch is not only difficult but requires a lot of computing power and time. 

To circumvent training from scratch of a several-billion-parameters model, we restrict ourselves to fine-tuning a pre-trained existing model. This has led us to select StarCoder LLM, based on GPT BigCode architecture \citep{allal2023santacoder}, which was designed for code generation and was fine-tuned specifically for Python PL. This was considered one of the best model when starting our research in code generation and thus was chosen as the backbone model.

Our approach is based on leveraging pre-trained StarCoder as the spine of our architecture while adding, \textit{on the side}, a contained number of new parameters that are used only when needed. Given that our final objective is to translate different input code languages to Python, a new set of additional parameters are created for each potential input language. Finally, the selection of which set of parameters to use along the backbone StarCoder is handled by a Mixture-of-Expert (MoE) \citep{shazeer2017outrageously} -inspired method that we describe below.

The introduction of task-specific parameters may bridge the performance gap between encoder-decoder and decoder-only models when considered in a restrained framework: we do not aim to build a general-purpose LLM but only a code translation model.

\subsection{StarCoder Backbone}
Decoder-only StarCoder has 15.5B parameters.  It is composed of 40 decoder blocks and has a context length of 8k tokens. It is a Python variant LLM since it has been fine-tuned from StarCoderBase on the Python subset of training data. It is designed as a code generator and is primarily able to generate programs and fill-in-the-middle code. On common evaluation datasets, it was ranked the best or among the bests against comparable and sometimes much bigger models.

We make the assumption that the model is excellent at producing Python outputs and can be fine-tuned to an any PL-to-Python translator. 

\subsection{LoRA Fine-tuning} 
We tried multiple reduced-parameters fine-tuning methods \citep{peft}  to contain the total number of parameters to train including the Low-Rank Adaptation (LoRA) \citep{hu2021lora} and Prefix-Tuning \citep{li2021prefixtuning}. We also tried combining them when they were orthogonal, but we observed that applying solely LoRA method yielded the best results. This observation is further explained in \cite{he2022unified}.

LoRA fine-tuning method shows best results and has a double advantage in reducing the parameters used: the targeted matrix' rank can be reduced by a huge factor, and it was observed that targeting only a small number of matrices in the attention mechanism works well. 

Assuming that the parameters update has an intrinsic rank $r$ much lower than its actual rank, we can rewrite it as $\Delta W = B \times A$ where $\Delta W \in \mathbb{R}^{d \times k}$, $A \in \mathbb{R}^{r \times k}$ and $B \in \mathbb{R}^{d \times r}$. $A$ and $B$ are the only trainable parameters, while $W$ remains frozen. $\Delta W$ is then scaled by $\frac{\alpha}{r}$, where $\alpha$ is a constant, and $r$ represents the intrinsic rank $r$.

\begin{figure}[h]
    \centering
    \includegraphics[scale=0.45]{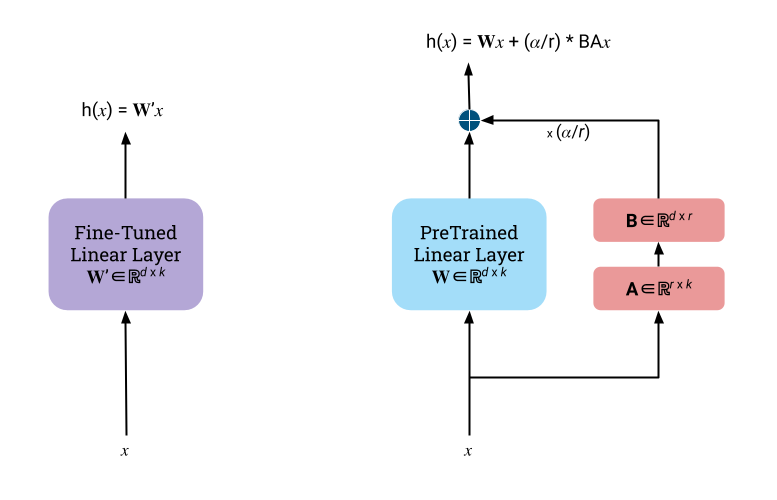}
    \caption{In inference, LoRA weights $A$ and $B$ can either be seamlessly merged by updating $W$ with $W' = W + \frac{\alpha}{r}\Delta W$ (left) or kept on a parallel route (right).}
    \label{LoRA}
\end{figure}

Moreover, LoRA has an additional interesting feature that it is not impacting the backbone architecture. During training, the LoRA method creates new matrices of reduced-rank and updates those instead of changing the original parameters. During inference, one can either merge these new matrices to the existing ones updating the model parameters, or they may keep them separated leaving the base model unchanged (i.e. the input is passed both through the original layer and the LoRA layer simultaneously, after which the results are added together) as this is depicted in Figure \ref{LoRA}.

Using this last feature and keeping untouched the StarCoder backbone while creating a contained number of parameters to train allows us to consider the following: creating one expert model for each PL to translate. Because we only have to store the backbone parameters once, we can fine-tune our model multiple times, one for each PL, creating a set of parameters every time and keep all these new parameters \textit{on the side}. During inference, we select only the right set of parameters to use along the StarCoder backbone. Moreover, adding additional language experts does not create memory loss since each language translation is independent, and the original model weights are preserved. This is also observed by \cite{gururangan2021demix}.

This architecture leverages both the precision of an expert model being trained on a specific task (e.g. translating C++ to Python) and the efficiency of training and using a small number of parameters.

\subsection{MoE Routing}
The main challenge of this architecture is to ensure that the model routes to the right expert. That is, given an input in X language, the model should use the X expert - "expert" refers to the whole set of additional parameters throughout the architecture that were trained for this PL.

To route towards the right set of parameters, we get inspiration from the MoE model. The MoE is originally designed to compute the top k experts to use for a given input. It was thought for a single task where the experts were simple MLPs all trained for the same task with similar data. In other words, MoE is originally thought to improve precision at relatively small inference cost. 

\begin{figure}[h]
    \centering
    \includegraphics[scale=0.5]{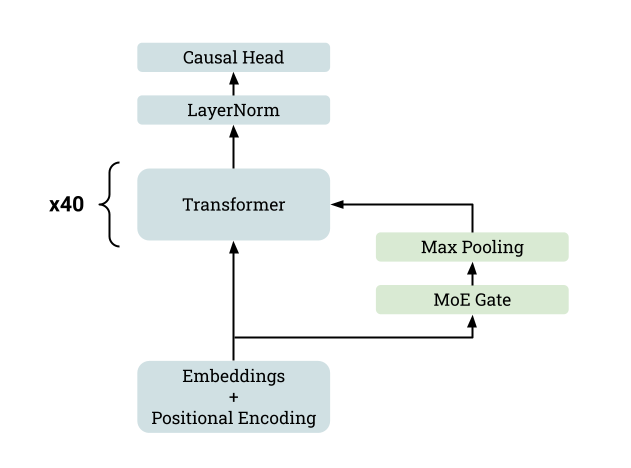}
    \caption{SteloCoder high-level architecture. The MoE gate's output corresponds to the expert to use (during inference) or the experts' weights (during training), which is injected in the transformer blocks where the LoRA experts are used accordingly.}
    \label{MoE}
\end{figure}

In our case, we adapt and simplify this idea while keeping the major advantage of MoE which is using small-sized experts in a greater context. As shown in figure \ref{MoE}, in parallel of the backbone model, we introduce a linear layer right after the embedding layer. The output's shape is \texttt{[batch\_size, sequence\_length, number\_experts]}. We aggregate the \texttt{sequence\_length} dimension to summarize the information from the input using a max pooling, resulting a shape of \texttt{[batch\_size, number\_experts]}. Applying a softmax, this yields a classification probability to determine which PL the input is. Finally, we compute the index associated with the maximum probability and propagate it throughout the transformer blocks to select the right expert.

\subsection{SteloCoder}
The final model SteloCoder is built in two steps: first we train individually the experts, and then we train the linear MoE gate to select the correct expert to use.

For the first phase, we consider one PL translation at a time. StarCoder is fine-tuned on a given PL translation using the LoRA method. LoRA is applied to three matrices in the attention mechanism (the linear layer producing the query, keys and values which is one matrix in BigCode implementation, the attention projection layer, and the MLP projection layer) within all 40 transformer blocks. The LoRA factor is set to $r=4$ (and $\alpha=32$). This sums up to an additional 9M parameters per expert. Once the model is trained for one PL-to-Python translation, the LoRA weights are not merged but rather stored and are used in parallel in the final model.

In the second phase, the experts are stored along the backbone model and are to be used only when needed. In this phase, we only train the MoE gate which is a linear layer of 31k parameters, everything else, including the expert parameters, are frozen. It is important to note that the argmax function (applied to determine the index of max-probability expert) is not differentiable. To circumvent this issue, we adopt slightly different behaviors in training and inference. During training, instead of using only the max-probability expert, we use the output of the softmax as weights. We pass the input through all the experts and calculate the weighted average of the outputs as our final expert output \citep{fedus2022switch}. This enables the backward propagation. Over training, the model encourages greater weight on the right expert, and during inference, only this one will be used.

SteloCoder is trained for translating five PLs to Python: C++, C\#, Java, JavaScript and PHP.

\section{Experimental Setup}

\subsection{Data For Model Training}

The data used for training, validation, and evaluation is sourced from XLCoST \citep{zhu2022xlcost}. 

Our initial step involves processing the dataset into JSON format, where each code sample is paired with its corresponding Python code piece. For the pre-processing, we employ a self-instruct method \citep{wang2023selfinstruct}. Before feeding the training data, we create a class to pad each sample to equal token length.

\subsubsection{Dataset}
XLCoST \citep{zhu2022xlcost}, the data source, is a comprehensive dataset for code generation and code retrieval tasks with multiple programming languages. Code generation tasks encompass code translation (code-to-code), code summarization (code-to-text), and code synthesis (text-to-code). Given SteloCoder's specific focus on code translation, we exclusively utilize data related to code translation for model training. Specifically, we use pair-wise code translation data in both snippet-level and program-level.

\subsubsection{Data Pre-processing}

\paragraph{JSON Format}

The data samples, organized in pairs, are initially stored in two separate files, with each file containing code data in a different PL. For example, in C++ to Python code translation data, all C++ code samples are stored in a \texttt{.cpp} file, and Python code samples correspondingly stored in a \texttt{.py} file. We first re-structure the data in JSON files to store code translation pairs. Each JSON element contains two key-value pairs, where the key specifies the name of the programming language, and the value stores the corresponding code snippet or program in that language. For instance, a code translation sample for cpp-to-python translation is preprocessed into the format: \texttt{"cpp" : cpp\_code, "py": py\_code}. 

For model training and validation, we utilize pair-wise translation data in both snippet-level and program-level due to CL strategy, which is discussed in the next section. Additionally, program-level data is used for evaluation. A summary of the dataset is presented in Table \ref{raw_data}.

\begin{table}[h]
  \centering
  \begin{tabular}{|c|c c|c c|c|} 
    \hline
    & \multicolumn{2}{c|}{\textbf{Train}} & \multicolumn{2}{c|}{\textbf{Validation}} & \multicolumn{1}{c|}{\textbf{Test}} \\
    \hline
    \textbf{Conversion} & \textbf{Snippet} & \textbf{Program} & \textbf{Snippet} & \textbf{Program} & \textbf{Program} \\
    \hline
    cpp-to-python & 80100 & 9139 & 3913 & 468 & 878 \\
    csharp-to-python & 75843 & 8826 & 3922 & 470 & 877 \\
    javascript-to-python & 67219 & 8182 & 3750 & 459 & 864 \\
    java-to-python & 77759 & 8991 & 3938 & 471 & 882 \\
    php-to-python & 17616 & 3003 & 923 & 153 & 304 \\
    c-to-python & 2478 & 380 & 311 & 59 & 48 \\
    \hline
  \end{tabular}
  \caption{Raw Data Summary}
  \label{raw_data}
\end{table}

Although C-to-Python translation data is available, its limited number of samples have a detrimental impact on the performance of fine-tuning and gating network training. Therefore, we excluded the C-to-Python translation functionality from our final model. 

\paragraph{Self-Instruct Data} To prompt the model with a supervised format \citep{wang2023selfinstruct}, we introduce tags indicating the PL name for each sample. Specifically, we added a tag token \texttt{<py>}, \texttt{<cpp>}, \texttt{<csharp>}, \texttt{<js>}, \texttt{<java>}, or \texttt{<php>} at the beginning of Python, C++, C\#, JavaScript, Java, or PHP code correspondingly.

\paragraph{Data Samples construction} To create the final dataset, we concatenate the two key-value pairs in each element from JSON files. Therefore, each data sample is in the format of: \texttt{<PL name> language\_code <py> python\_code}. An example below shows a snippet-level Java-to-Python sample in the training dataset.

\begin{figure}[h]
    \centering
    \includegraphics[scale=0.35]{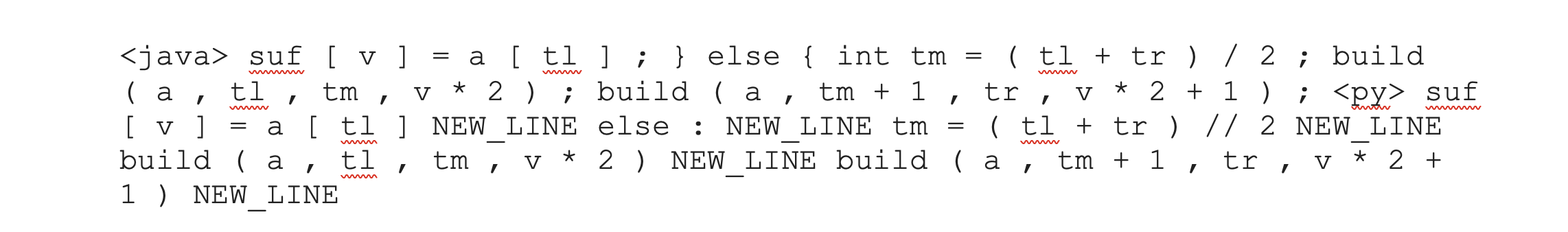}
    \caption{A training data sample for Java to Python code translation}
    \label{Data Example}
\end{figure}

\subsubsection{Data Split for CL Strategy}

While training SteloCoder, we adapt the CL strategy \citep{10.1145/1553374.1553380} \citep{Kocmi_2017} \citep{zhao2020reinforced}. CL proposes to feed the model with examples of increasing complexity during training which could help finding a better local minima and accelerate the convergence. To implement this strategy, we train the original StarCoder first in snippet-level data and then in program-level data. 

In open-source code translation data from XLCoST, snippet-level data is extracted from program-level code. In order to remove duplication, we partition the data based on token counts. Specifically, for program-level data, we calculate the total number of tokens in the training set and divide it in half. The resulting second half became the program-level training data, while the corresponding snippet-level data from the first half is designated as the snippet-level training data.

\subsubsection{Data Padding}

For efficient batch training, we standardize the token length of each sample by padding. To mitigate potential issues caused by few exceedingly long translation samples, which could lead to inefficient training due to extensive padding, we select the shortest 95\% of samples in terms of token count and pad them to a uniform length.

\subsubsection{Dataset for MoE Training}

To train the gating network in MoE framework, we curate a balanced dataset containing examples from all languages. Since C-to-Python translation was removed from the expert languages, we select the smallest program-level dataset size among the remaining five languages as the sample size for each PL translation data. Specifically, we chose the first 3003 samples from each dataset and combined them to form the MoE training dataset, which comprised 15,015 training samples.

\subsection{Evaluation Method}

We employ the CodeBLEU metric \citep{ren2020codebleu} as our evaluation criterion. CodeBLEU is a specialized metric for assessing code translation models. Apart from n-gram (BLEU) \citep{papineni-etal-2002-bleu} and weighted n-gram match, CodeBLEU also introduces syntactic match via matching AST trees and semantic data-flow match. The score can be interpreted as a percentage, with a balanced-weighted addition of BLEU, weighted BLEU, syntax match, and data-flow match. 

\subsection{Experimental Details}

SteloCoder underwent a two-phase training process, first acquiring the LoRA experts, one for each PL translation, and then obtaining weights for the gating network.

\subsubsection{Expert Acquisition}

\paragraph{LoRA Method for Fine-tuning} We employ the LoRA method for fine-tuning StarCoder. This approach enable us to save memory by retaining only low-rank matrices. In our implementation, the dimension of the low-rank matrices is set to 4, LoRA alpha was chosen as 32, and a LoRA dropout of 0.05 was applied. From observation, these hyperparameters chosen above yield good fine-tuning performance.

\paragraph{Fine-tuning with CL Strategy} As previously mentioned, we adopt a CL strategy to fine-tune the original StarCoder model. The strategy involved gradually increasing the complexity of training by starting with code snippet-level data for 2 epochs and subsequently changing to code program-level data for 2 epochs. In each epoch, we shuffled the data with different seeds. The learning rate varied between e-5 and e-6, with a decreasing learning rate at each stage.

\subsubsection{Gating Network Training}

From observation, factors that caused unstable training included inappropriate learning rate and unshuffled dataset. For final gating network training. Learning rate for gating network training was set as 5e-5, and the data was shuffled each time with different seeds.

\section{Results and Analysis}

\subsection{Performance after Fine-tuning}

We obtain expert weights from LoRA weights during the fine-tuning stage. Specifically, the LoRA weights are obtained from fine-tuning the original StarCoder on five distinct PLs-to-Python for code translation. Table \ref{experts_perf} demonstrates that each fine-tuned model outperformed the top performer on the XLCoST leaderboard in all languages except PHP. Each model performance surpasses the CodeBLEU score from the top performer in XLCoST leaderboard by at least 3 in C++, C\#, JavaScript, and Java, and PHP-to-Python performance is lagged by 1. As a result, each fine-tuned model can successfully translate from a designated PL to Python, including but not limit to package imports, Python-specific syntax, and the main function.

\begin{table}[ht]
  \centering
  \begin{tabular}{@{}ccc@{}} 
    \toprule
    & \textbf{CodeBLEU (single expert)} & \textbf{CodeBLEU (1st in XLCoST)} \\
    \midrule
    cpp-to-python & \underline{76.62} & 71.56 \\
    csharp-to-python & \underline{73.93} & 69.52 \\
    javascript-to-python & \underline{72.13} & 68.42  \\
    java-to-python & \underline{73.59} & 69.57  \\
    php-to-python & 71.21 & \underline{72.26}  \\
    \bottomrule
  \end{tabular}
  \caption{Performance for single expert}
  \label{experts_perf}
\end{table}

\subsection{Expert Routing Performance}

After incorporating experts into SteloCoder, we trained SteloCoder to recognize which PL the input belongs to. Table \ref{confus_matrix} illustrates the model's ability to identify C++, PHP, and JavaScript with close to 95\% accuracy. C\# and Java proved more challenging as they have very close syntaxes, resulting in an accuracy of 67\%. This is not an issue as we expect the C\# and Java experts to have roughly similar behaviors for this exact same reason.

\begin{table}[ht]
\centering
\begin{tabular}{cccccc}
           & C++ & PHP & JavaScript & C\# & Java \\ \cline{2-6} 
C++        & 149 & 0   & 1          & 0   & 3    \\
PHP        & 1   & 152 & 0          & 0   & 0    \\
JavaScript & 1   & 0   & 145        & 0   & 0    \\
C\#        & 2   & 1   & 7          & 150 & 47   \\
Java       & 0   & 0   & 0          & 3   & 103  \\ \toprule
Total      & 153 & 153 & 153        & 153 & 153 
\end{tabular}
\caption{Test Confusion Matrix. The predictions are on the left (rows), the true labels are above (columns).}
\label{confus_matrix}
\end{table}

\subsection{SteloCoder Performance}

We now test SteloCoder on test cases from XLCoST leaderboard and display the results in Table \ref{results}. We achieve scores higher than current best scores in XLCoST leaderboard except PHP. We interpret this improvement as two sided: on the one hand we have trained specialized expert for each specific task, on the other we believe that StarCoder's original ability to produce qualitative Python code made such high scores possible.

We also compare the final model output against the original single-expert fine-tuned model outputs and observe that the routing network do not impact the performance at all. In some cases we even observed a very slight improvement. This is expected as the final model has more degrees of freedom and can choose different experts (for instance, the C\# expert may output better code quality than the Java expert, and ends up being regularly chosen for Java inputs).

\begin{table}[h]
  \centering
  \begin{tabular}{@{}ccc@{}} 
    \toprule
    & \textbf{CodeBLEU (SteloCoder)} & \textbf{CodeBLEU (1st in XLCoST)} \\
    \midrule
    cpp-to-python & \underline{75.42} & 71.56 \\
    csharp-to-python & \underline{74.83} & 69.52 \\
    javascript-to-python & \underline{73.05} & 68.42  \\
    java-to-python & \underline{74.39} & 69.57  \\
    php-to-python & 71.11 & \underline{72.26}  \\
    \bottomrule
  \end{tabular}
  \caption{Performance for SteloCoder}
  \label{results}
\end{table}

Moreover, we tested SteloCoder's performance when the input PL was not specified. That is, we replaced the language token \texttt{<PL name>} by a generic token \texttt{<code>}. The model showed no sign of difficulty and was still able to detect the PL used (as shown in Appendix \ref{ssec:num1}). SteloCoder consistently selected the correct experts with high probabilities. 

\section{Discussion}

SteloCoder yields high score in codeBLEU metric. As mentioned before, codeBLEU score is a weighted combination of BLEU, weighted BLEU, syntax match, and data-flow match. While codeBLEU exhibits a stronger correlation with code quality compared to relying solely on the BLEU score, the inclusion of the BLEU score in the codeBLEU metric introduces the possibility that translated codes achieving high BLEU scores may not necessarily excel in real-world performance. This is similar to a phenomenon observed in \cite{radford2023robust}. It is still an open question on how to design a better metric which better aligns with human judgement.

SteloCoder presents great performance. However, it is at the cost of flexibility. Given its task-specific architecture, modifying the model can be relatively complex. For instance, adding a PL to translate would require training a new set of experts and re-training the MoE gate from scratch as its shape would change. However, this process is low time-cost since only the gating MoE gate needs to be re-trained.

The final LoRA logistics chosen here - keeping an unmerged parallel route - is not computationally optimal during inference. This design requires three matrix multiplications when passing through a LoRA-decomposed matrix. As shown in figure \ref{LoRA}, the input is multiplied with the original weights matrix $W$ and is also simultaneously multiplied with LoRA matrices $A$ and then $B$. A computationally efficient alternative is to introduce the new matrix $C = B \times A$, computed only once. Whenever an input is going through a LoRA-decomposed matrix, one can first merge $C$ in $W$ and perform the matrix multiplication. Afterwards, the $C$ matrix is unmerged from $W'$ using a simple subtraction. This requires only one matrix multiplication and is achievable at the (very limited) cost of memory efficiency.

Moreover, other models have been showing great results in code translation while maintaining a good understanding and generation in other tasks \citep{wang2021codet5} \citep{feng2020codebert} \citep{ahmad2021unified}. Even though SteloCoder is still able to perform Python code generation tasks (as shown in Appendix \ref{ssec:num2}), it remains highly limited, and it does not benefit from any broader knowledge as StarCoder was already a code-specific model. The recent release of Code Llama, and especially Code Llama - Python \citep{rozière2023code}, encourages us to think that a similar task-specific architecture applied to a PLM fine-tuned from a general-purpose LLM (such as Code LLaMA or Code LLaMA - Python) could yield excellent results while offering a broader general knowledge. This is not negligible as reasoning learned through general data could help design algorithms and conversely the analysis of an unknown program could be done via the use of external knowledge.

\section{Conclusion}

We present SteloCoder, a decoder-only PLM for five PL-to-Python code translation with only 45M more parameters than the backbone. Its performance surpasses other models in XLCoST leaderboard in PL-to-Python code translation with an average of 3.5 codeBLEU score. SteloCoder has been efficiently trained for 32 hours on one 80Gb A100 device. The main characteristic of SteloCoder is that it was built with several task-specific experts that are used only when needed. We designed this architecture leveraging the LoRA and MoE designs. We believe this unlocks a better precision while remaining computationally and memory efficient during inference.

\bibliography{iclr2023_conference}
\bibliographystyle{iclr2023_conference}

\newpage
\section{Appendix}
\subsection{Code Translation Examples}
\label{ssec:num1}
We give below, in figures \ref{tran_cpp} \ref{tran_php} \ref{tran_java}, a few examples of translations performed by our model from an unlabeled PL (\texttt{<PL name>} token is replaced by \texttt{<code>}.
The inputs and outputs are cleaned up manually for a better visual understanding (spaces, \texttt{<code>}, \texttt{<py>} and \texttt{<|end-of-text|>} tokens removed, and \texttt{NEWLINE}, \texttt{INDENT}, \texttt{DEDENT} tokens replaced).
\begin{figure}[hbtp]
    \begin{lstlisting}[language=c,frame=single]
#include "bits/stdc++.h"
using namespace std;
bool isPerfectSquare(long double x) {
    long double sr = sqrt(x); 
    return ((sr - floor(sr)) == 0); 
} 
void checkSunnyNumber(int N) {
    if (isPerfectSquare(N + 1)) { 
        cout << "Yes STRNEWLINE"; 
    } 
    else { 
        cout << "No STRNEWLINE"; 
    } 
}
int main() {int N = 8 ; checkSunnyNumber(N); return 0;}
    \end{lstlisting}
    \begin{lstlisting}[language=python,frame=single]
from math import sqrt, floor
def isPerfectSquare(x):
    sr = sqrt(x)
    return ((sr - floor(sr)) == 0)
def checkSunnyNumber(N):
    if (isPerfectSquare(N + 1)):
        print("Yes")
    else:
        print("No")
if __name__ == '__main__':
    N = 8
    checkSunnyNumber(N)
    \end{lstlisting}
    \caption{In the upper box: C++ input code. In the lower box: Python output code.}
    \label{tran_cpp}
\end{figure}

\begin{figure}[hbtp]
    \begin{lstlisting}[language=php,frame=single]
<?php
function possibleToReach($a,$b) {
    $c = $a*$b;
    $re1 = $a/$c;
    $re2 = $b/$c;
    
    if (($re1*$re1*$re2 == $a) && ($re2*$re2*$re1 == $b)) {
        return 1;
    } else {
        return -1;
    }
}
$A = 60;
$B = 450;
if (possibleToReach($A,$B)) {
    echo "yes";
} else {
    echo "no";
}
?>
    \end{lstlisting}
    \begin{lstlisting}[language=python,frame=single]
def possibleToReach(a,b):
    c = (a*b)
    re1 = a/c 
    re2 = b/c
    if ((re1*re1*re2 == a) and (re2*re2*re1 == b)):
        return 1
    else:
        return -1
A = 60
B = 450
if (possibleToReach(A,B)):
    print("yes")
else: 
    print("no")
    \end{lstlisting}
    \caption{In the upper box: PHP input code. In the lower box: Python output code.}
    \label{tran_php}
\end{figure}

\begin{figure}[hbtp]
    \begin{lstlisting}[language=java,frame=single]
class GFG {
    static final int mod = (int) (1e9 + 7);
    static int ways(int arr[], int n) {
        int dp[] = new int[n + 1];
        dp[n - 1] = 1;
        for (int i = n - 2; i >= 0; i--) {
            dp[i] = 0;
            for (int j = 1; ((j + i) < n && j <= arr[i]); j++) {
                dp[i] += dp[i + j];
                dp[i] %= mod;
            }
        }
        return dp[0] % mod;
    }
    public static void main(String[] args) {
        int arr[] = {5, 3, 1, 4, 3};
        int n = arr.length;
        System.out.println(ways(arr, n) % mod);
    }
}

    \end{lstlisting}
    \begin{lstlisting}[language=python,frame=single]
mod = 1000000007
def ways(arr,n):
    dp = [0]*(n + 1)
    dp[n - 1] = 1 
    for i in range(n - 2,-1,-1):
        dp[i] = 0
        for j in range(1,min(n - i,arr[i]) + 1):
            dp[i] += dp[i + j]
            dp[i] %= mod
    return dp[0] % mod 
if __name__ == '__main__':
    arr = [5,3,1,4,3]
    n = len(arr) 
    print(ways(arr,n) % mod)
    \end{lstlisting}
    \caption{In the upper box: Java input code. In the lower box: Python output code.}
    \label{tran_java}
\end{figure}

\subsection{Code Generation Example} \label{ssec:num2}
Below is an example, in figure \ref{gen}, of code generation performed by SteloCoder on a simple example.
The input and output are cleaned up manually for a better visual understanding (spaces, \texttt{<code>}, \texttt{<py>} and \texttt{<|end-of-text|>} tokens removed, and \texttt{NEWLINE}, \texttt{INDENT}, \texttt{DEDENT} tokens replaced).
\begin{figure}[hbtp]
    \begin{lstlisting}[language=,frame=single]
Give code to compute the first prime numbers.
    \end{lstlisting}
    \begin{lstlisting}[language=python,frame=single]
def prime(n): 
    if n == 1: 
        return False 
    if n == 2: 
        return True 
    if n%2 == 0 : 
        return False 
    for i in range(3,int(n**0.5)+1,2): 
        if n%i == 0: 
            return False 
    return True
n = int(input()) 
for i in range(1,n+1): 
    if prime(i): 
        print(i)
    \end{lstlisting}
    \caption{In the upper box: input in natural language. In the lower box: Python output code.}
    \label{gen}
\end{figure}

\end{document}